\pdfoutput=1

\documentclass[11pt,table]{article}

\usepackage{acl}

\usepackage{times}
\usepackage{latexsym}
\usepackage{lipsum}  
\usepackage{graphicx}
\usepackage{stfloats}
\usepackage{multirow}
\usepackage{booktabs,makecell}
\usepackage{xcolor,caption}

\usepackage{silence}
\usepackage[T1]{fontenc}

\usepackage[utf8]{inputenc}

\usepackage{microtype}

\title{Leveraging supplementary text data to kick-start automatic speech recognition system development with limited transcriptions}

\author{Nay San\textsuperscript{1}, Martijn Bartelds\textsuperscript{2}, Blaine Billings\textsuperscript{3}, Ella de Falco\textsuperscript{4}, Hendi Feriza, \\
\textbf{ Johan Safri, Wawan Sahrozi, Ben Foley\textsuperscript{5}, Bradley McDonnell\textsuperscript{3}, Dan Jurafsky\textsuperscript{1}} \\
\textsuperscript{1}Stanford University; \textsuperscript{2}University of Groningen; \textsuperscript{3}University of Hawai'i at M\=anoa; \\
  \textsuperscript{3}University of Washington; \textsuperscript{5}University of Queensland
  \\
  \texttt{nay.san@stanford.edu}
}

\begin{document}
\maketitle

\begin{abstract}
Recent research using pre-trained transformer models suggests that just 10 minutes of transcribed speech may be enough to fine-tune such a model for automatic speech recognition (ASR) --- at least if we can also leverage vast amounts of text data (803 million tokens).~But is that much text data necessary?~We study the use of different amounts of text data, both for creating a lexicon that constrains ASR decoding to possible words (e.g. *\textit{dogz} vs. \textit{dogs}), and for training larger language models that bias the system toward probable word sequences (e.g. \textit{too dogs} vs. \textit{two dogs}).~We perform experiments using 10 minutes of transcribed speech from English (for replicating prior work) and two additional pairs of languages differing in the availability of supplemental text data: Gronings and Frisian (\textasciitilde{}7.5M token corpora available), and Besemah and Nasal (only small lexica available).~For all languages, we found that using only a lexicon did not appreciably improve ASR performance.~For Gronings and Frisian, we found that lexica and language models derived from `novel-length' 80k token subcorpora reduced the word error rate (WER) to 39\% on average.~Our findings suggest that where a text corpus in the upper tens of thousands of tokens or more is available, fine-tuning a transformer model with just tens of minutes of transcribed speech holds some promise towards obtaining human-correctable transcriptions near the 30\% WER rule-of-thumb.

\end{abstract}

\section{Introduction}

Automatic speech recognition (ASR) systems can help speed up the labour-intensive process of transcribing speech by providing human-correctable first-pass transcriptions.
The Catch-22, however, is that some speech must first be manually transcribed in order to develop a sufficiently performant ASR system.
As such, recently developed pre-trained transformer models for speech such as wav2vec 2.0 \cite{NEURIPS2020_92d1e1eb} have spurred on much experimentation to leverage them to accelerate transcription workflows in language documentation contexts \citep[e.g.][]{coto-solano-etal-2022-development,guillaume2022fine,macaire-etal-2022-automatic,san-etal-2022-automated,zhang-etal-2022-nlp}, given a much reduced upfront cost in terms of transcribed speech.

Impressively, \citet{NEURIPS2020_92d1e1eb} showed that a wav2vec 2.0 model pre-trained on 960 hours of untranscribed English speech required just 10 minutes of transcriptions to yield competitive results on the LibriSpeech ASR benchmark \cite{panayotov2015librispeech}.
This result, however, also leveraged the official LibriSpeech lexicon and language model derived from the entire 803 million token text corpus based on 14.5k public domain books.\footnote{\url{https://www.openslr.org/11/}}
However, an in-domain text corpus of such size is not within immediate reach for many other languages.
In this paper, we investigate the real-world ASR performance achievable with 10 minutes of transcribed speech along with more tenable amounts of supplemental text data.

We first replicated the wav2vec 2.0 experiments on LibriSpeech and then extended them by creating lexica and language models using reduced amounts of supplemental text data (8M, 80k, and 8k tokens) from both the in-domain LibriSpeech text corpus and an out-of-domain corpus composed of web-scraped text \citep[Common Crawl:][]{Buck-commoncrawl}.
Our experiments showed that first-pass transcriptions within the 20-30\% word error rate (WER) rule-of-thumb \cite{gaur2016effects,sperber-etal-2016-optimizing} can indeed be obtained with just 10 minutes of transcribed speech if supplemented with a lexicon and language model derived from a corpus with at least 80k tokens.


We then performed analogous experiments with two pairs of languages differing in the availability of supplemental texts: Gronings and Frisian (two Germanic languages for which a modest amount of external text-only data can be sourced), and Besemah and Nasal (two Malayo-Polynesian languages for which the documentation projects' own materials constitute the only available text data).
For these experiments, we used 10 minutes of audio from each of the languages to fine-tune the multilingual wav2vec 2.0 XLS-R model \cite{babu2021xls} that is pre-trained on 450k hours of speech from 128 languages.

For all four languages we found that using only a lexicon to restrict ASR system output to possible words did not appreciably reduce the WER.
For Gronings and Frisian, where there was sufficient text data to derive both lexica and language models from various samples (\textasciitilde{}7.5M, 80k, and 8k tokens), we found that those derived from 80k tokens of text reduced the WER to 34.9\% for Gronings and 43.0\% for Frisian (mean: 39.0\%).

While these error rates are above the 20-30\% rule-of-thumb for first-pass transcriptions \citep{gaur2016effects,Sperber2017b}, it is worth noting that the multilingual XLS-R model used for Gronings and Frisian was not pre-trained on the target datasets.
This is in contrast to the English wav2vec 2.0 model, which was pre-trained on all 960 hours of the target LibriSpeech dataset.
Thus, if combined with a domain adaptation technique \cite[e.g.~continued pretraining:][]{gururangan2020don}, our results suggest that where 80k or more of supplementary text is available, fine-tuning a pre-trained model with just tens of minutes of transcribed speech could help kick-start a virtuous cycle of data collection and training for ASR system development.

\section{Motivations}

\subsection{Related work}



As mentioned above, there have been several recent studies appraising the utility of pre-trained transformers for ASR in language documentation settings.
\citet{coto-solano-2022-evaluating} reported that the \mbox{XLS-R} model fine-tuned with 4 hours of transcribed speech from Cook Islands M\=aori yielded a WER of 22.9\% without using a language model (LM).
Similarly, \citet{guillaume2022fine} found that the XLS-R model fine-tuned with 10 hours of transcriptions from Japhug yielded a WER of 18.5\% also without a LM.
Notably, both these studies explicitly mentioned that the system outputs (\textasciitilde{}20\% WER) did indeed appear suitable as first-pass transcriptions for their respective projects and that LM integration is a clear next step.

Of the two studies that have examined the use of a LM with a fine-tuned wav2vec 2.0 model for language documentation projects \citep{macaire-etal-2022-automatic,san-etal-2022-automated}, a common theme has been to examine the effect of varying amounts of transcribed speech (e.g.~10--70 minutes) with and without the use of a LM trained on the full corpus (e.g.~74.5k tokens of text).
Similarly, the original wav2vec 2.0 experiments by \citet{NEURIPS2020_92d1e1eb} also only examined the use of different amounts of fine-tuning data (10 minutes to 960 hours), with or without the use of LMs trained on an 803M token corpus.

Given the broader community interest in using fine-tuned wav2vec 2.0 models with LM integration, we undertook a series of experiments holding constant the amount of fine-tuning data (i.e.~10 minutes) while varying the amount of LM training text (e.g.~8k--8M tokens) to complement the aforementioned studies.

\subsection{Language projects}

For Besemah, Nasal, and Gronings, the motivation for the development of ASR systems is to help derive first-pass transcriptions and hence accelerate the process of indexing a large collection of audiovisual materials.
Besemah (ISO 639-3: pse) and Nasal (ISO 639-3: nsy) are two Malayo-Polynesian languages spoken in Sumatra, Indonesia.
For both languages, approximately 45 hours of informal conversations have been collected as part of fieldwork by author BM.
Part of these collections have been transcribed by author BM and collaborators from the Besemah (author HF) and Nasal communities (authors JS and WS).
The collections are managed by authors BB and EF and are accessible at \mbox{PARADISEC} \citep{mcdonnell2008besemah,mcdonnell2019nasal}.

Gronings (ISO 639-3: gos) is a Low Saxon language variant spoken in the province of Groningen in the Netherlands.
An ongoing language documentation project (of which author MB is part) has so far recorded approximately 15 hours of speech, with more being continually gathered.
The materials will be used to create ASR and text-to-speech systems to be made available through an online cultural portal.\footnote{\url{https://www.woordwaark.nl}}~For an additional point of comparison in our experiments, we also included West Frisian (another minority language spoken in the Netherlands; ISO 639-3: fry), using data from the FAME! ASR corpus \citep{yilmaz2017longitudinal}.

\section{Method}

For English, we used the `\texttt{1h/0}' ten minute training set defined in Libri-Light \citep{kahn2020libri} to approximate the fine-tuning experiments in \citet{NEURIPS2020_92d1e1eb}.~As supplementary text data, we used the official normalised LibriSpeech as the in-domain corpus and Common Crawl \citep{Buck-commoncrawl} as the out-of-domain corpus.

For Besemah, Nasal, Gronings and Frisian, we constructed comparable 4-hour datasets for our broader set of ASR experiments.~Each dataset is composed of an 80/10/10 train/dev/test split, approximately 24 minutes for each of the dev and test sets and 3.2 hours for training set, from which we sampled 10 minutes for the experiments reported here.~As supplementary text data, we sourced two 9.5k word lists for Nasal and Besemah from project materials and remaining transcriptions not included in the ASR corpora.~For Gronings and Frisian, we sourced two \textasciitilde{}7.5M corpora used in \citet{de-vries-etal-2021-adapting}.

For English, we used the monolingual English wav2vec 2.0 model \citep{NEURIPS2020_92d1e1eb}.~For the other four languages, we used the multilingual XLS-R model \citep{babu2021xls}.~For fine-tuning these models, we used the HuggingFace Transformers library \citep{wolf2019huggingface}.
For beam search decoding with a lexicon and language model, we used the torchaudio implementation of the Flashlight decoder \citep[formerly wav2letter:][]{kahn2022flashlight} used in the original paper.
We fine-tuned each model for 12k steps and then selected the best checkpoint based on the dev set WER using greedy decoding (for Besemah and Nasal) or beam search decoding with fixed parameters (language model weight: 2, word insertion penalty: -1).
We then performed a parameter search with the best checkpoint to further optimise the dev set WER.
We repeated the decoding experiments with 5 different random samples of each size of sub-corpora.
All our experiment code is available on GitHub.\footnote{\url{https://anonymous.4open.science/r/w2v2-10min-exps-computel6}}

\section{Results and discussion}
\label{sec:resdisc}

The results from our experiments are collated in Table \ref{tab:results}.
For English, we found that fine-tuning with 10 minutes of transcribed speech yielded a test set WER of 40.2 (Row E1) without the use of supplementary texts and a WER of 14.2 (Row E3) using a lexicon and language model derived from the full 803M LibriSpeech text corpus.~These WERs are consistent with those reported by \citet[][Table 9]{NEURIPS2020_92d1e1eb}, respectively: 45.3 and 13.1, allowing for some error likely from differences in the 10 minute samples.\footnote{There was no indication as to which of the six 10-minute Libri-Light training sets were used in the original experiments.}

For all languages, we found that using only a lexicon did not appreciably reduce the WER.
For example, for English, the test set WER remained practically the same with or without a lexicon (Row~E1~vs.~E2:~40.2~vs.~40.5).
For Nasal, the test set WER increased from 70.7 without a lexicon (Row N1) to 75.1 when using a small 9.5k token lexicon (Row N2), which led to many erroneous substitutions from the combination of an already high WER and high out-of-vocabulary rate.

For English, Gronings, and Frisian, we found that using lexica and language models derived from 80k of out-of-domain supplementary texts appreciably reduced the mean test set WER.
For English, the test set WER was reduced from 40.2 (Row E1) to 27.7 (Row E8).
For Gronings, the test set WER was reduced from 44.0 (Row G1) to 34.9 (Row G4).
For Frisian, the test set WER was reduced from 53.1 (Row F1) to 43.0 (Row F4).
These results suggest that where supplemental texts in the upper tens of thousands or more are available, fine-tuning a wav2vec 2.0 model with just tens of minutes of speech holds promise for deriving first-pass transcriptions near the 20-30\% rule-of-thumb.

For English, the acceptably low sub-30\% WER likely reflect an idealised set of circumstances in that the wav2vec 2.0 model used was pre-trained on the target dataset and that the genre of the audio is read speech.
Regarding genre, the higher proportion of read speech in the Gronings dataset likely reflects its lower WER compared to Frisian, which has mainly news and radio broadcasts.
Regarding pre-training, the multilingual XLS-R model used for Gronings and Frisian was not pre-trained on these datasets.~This disadvantage could be partially overcome by a domain adaptation technique such as continued pre-training \citep{gururangan2020don} or by using a wav2vec 2.0 model pre-trained on a similar language such as Dutch \citep{bartelds-wieling-2022-quantifying}.
\citet{macaire-etal-2022-automatic} report a 5--9\% reduction in WER when using a wav2vec 2.0 model pre-trained on French over a multilingual model when fine-tuning for ASR on Gwadloupéyen and Morisien, two French-based creole languages.

%
    \renewcommand{\topfraction}{0.99}	
    \renewcommand{\bottomfraction}{0.99}	
    \setcounter{topnumber}{2}
    \setcounter{bottomnumber}{2}
    \setcounter{totalnumber}{4}     
    \setcounter{dbltopnumber}{2}    
    \renewcommand{\dbltopfraction}{0.9}	
    \renewcommand{\textfraction}{0.07}	
    \renewcommand{\floatpagefraction}{0.7}	
    \renewcommand{\dblfloatpagefraction}{0.7}	


\begin{figure*}[!ht]
\centering
\scalebox{0.83}{
\begin{tabular}{llllll}
\toprule
\multirow{2}{*}[-0.2em]{\makecell[l]{wav2vec 2.0 model\\(Fine-tuning data, 10 mins)}} & \multirow{2}{*}[-0.2em]{\makecell[l]{Language model\\(Text corpus size, no. of tokens)}} & \multicolumn{2}{c}{Mean WER (SD)} & \multicolumn{2}{c}{Mean CER (SD)} \\ \cmidrule(lr){3-4} \cmidrule(lr){5-6} 

&                                                   & \makecell[c]{dev}        & \makecell[c]{test}       & \makecell[c]{dev}        & \makecell[c]{test}       \\ \midrule\midrule

                                

\multirow{10}{*}[-2em]{\makecell[l]{LS-960 (English)}} 
& E1. None                       & 40.4 & 40.2 & 13.0 & 12.8 \\ \cmidrule{2-6}
& E2. None, lexicon only (973k)  & 40.1 & 40.5 & 12.3 & 12.2 \\ \cmidrule{2-6}
& \multicolumn{5}{c}{\textsc{LibriSpeech} (in-domain)}                     \\[0.25em]
& E3. 4-gram, full corpus (803M)           & 13.4 & 14.2 & 6.51 & 6.75 \\ \cmidrule{2-6}
& E4. 4-gram, subset (8M)                   & 15.9 (0.09) & 16.6 (0.07) & 7.41 (0.05) & 7.58 (0.07) \\ \cmidrule{2-6}
& E5. 3-gram, subset (80k)                   & 23.5 (0.13) & 24.1 (0.12) & 10.5 (0.07) & 10.6 (0.10) \\ \cmidrule{2-6}
& E6. 3-gram, subset (\textbf{8k})           & \textbf{36.5} (0.20) & \textbf{36.8} (0.34) & 17.1 (0.09) & 17.0 (0.15) \\ \cmidrule{2-6}
& \multicolumn{5}{c}{\textsc{CommonCrawl-En} (out-of-domain)}                     \\[0.25em]
& E8. 3-gram, subset (\textbf{80k})       & \textbf{27.7} (0.13) & \textbf{27.7} (0.10) & 12.1 (0.13) & 12.0 (0.13) \\ \cmidrule{2-6}
& E9. 3-gram, subset (8k)       & 39.8 (0.12) & 39.8 (0.24) & 18.5 (0.19) & 18.3 (0.22) \\ \midrule

\multirow{5}{*}[-1em]{XLS-R (Gronings)}
& G1. None                                   & 43.8 & 44.0 & 11.2 & 11.3 \\ \cmidrule{2-6}
& G2. None, lexicon only (216k)          & 41.4 & 40.7 & 10.5 & 10.3 \\ \cmidrule{2-6}
& G3. 4-gram, full corpus (7.6M)              & 23.4 (0.22) & 22.3 (0.22) & 7.61 (0.62) & 7.44 (0.71) \\ \cmidrule{2-6}
& G4. 3-gram, subset (\textbf{80k})                 & \textbf{35.6} (0.30) & \textbf{34.9} (0.51) & 11.8 (0.28) & 11.4 (0.34)	 \\ \cmidrule{2-6}
& G5. 3-gram, subset (8k)                & 45.6 (0.41) & 46.3 (0.24) & 16.9 (0.67) & 16.7 (0.43) \\ \midrule

\multirow{5}{*}[-1em]{XLS-R (Frisian)}
& F1. None                                     & 55.1 & 53.1 & 18.6 & 18.3 \\ \cmidrule{2-6}
& F2. None, lexicon only (251k)                & 53.8 & 51.7 & 17.9 & 17.6 \\ \cmidrule{2-6}
& F3. 4-gram, full corpus (7.4M)               & 36.7 & 35.2 & 16.3 & 15.9 \\ \cmidrule{2-6}
& F4. 3-gram, subset (\textbf{80k})            & \textbf{44.4} (0.37) & \textbf{43.0} (0.22) & 20.3 (0.16) & 19.9 (0.22) \\ \cmidrule{2-6}
& F5. 3-gram, subset (8k)                 & 54.2 (0.39) & 52.2 (0.29) & 26.2 (0.30) & 26.0 (0.25) \\ \midrule

\multirow{2}{*}{XLS-R (Besemah)}
& B1. None                                & 62.3 & 62.1 & 21.4 & 21.2 \\       \cmidrule{2-6}
& B2. None, lexicon only (9.5k)           & 59.9 & 62.2 & 20.1 & 21.2 \\ \midrule

\multirow{2}{*}{XLS-R (Nasal)} & N1. None                  & 67.2 & 70.7 & 23.1 & 25.5 \\ \cmidrule{2-6}
& N2. None, lexicon only (9.5k)            & 70.1 & 75.1 & 24.1 & 26.1 \\ 
\bottomrule

\end{tabular}}
\captionof{table}{Results of fine-tuning wav2vec 2.0 models (LS960: pre-trained on 960 hours of English; XLS-R pre-trained on 400k hours of speech from 128 languages) for  automatic speech recognition (ASR) using only 10 minutes of transcribed speech from a target language and with varying amounts of reliance on supplementary text data. ASR performance measured by word error rate (WER) and character error rate (CER). For subset experiments, reported means and standard deviations (in parentheses) were derived across 5 runs with different samples of the text corpus.
Alphanumeric labels (E1--N1) used to assist with in-text references to results. Highlighted numbers indicate the smallest supplementary text corpus size which yielded an appreciably lower WER.}
\label{tab:results}
\end{figure*}

For languages where tens of thousands of tokens in supplementary texts are not available, we suspect it remains unavoidable for the time being that more than 10 minutes of transcribed speech is required to begin ASR system development.~For reducing the out-of-vocabulary rate with lexica and LMs derived from small text corpora, decomposing words into sub-word units (e.g. syllables) may be worth investigating.
In trialling an interactive transcription app for Kunwinjku, \citet{le-ferrand-etal-2022-fashioning} report that incorporating syllable unigram frequencies derived from a word list improved the word retrieval performance by 4\% (F-score).

\section{Conclusion}

We investigated the real-world performance obtainable when leveraging various amounts of supplemental text data to help kick-start automatic speech recognition (ASR) system development with only a limited amount of transcribed speech.~Our results suggest that fine-tuning a pre-trained transformer model with just 10 minutes of transcribed speech may hold some promise for deriving human-correctable first-pass transcriptions if the ASR system can incorporate a lexicon and language model derived from a `novel-length' corpus with at least 80,000 tokens or more of text.

\section*{Acknowledgements}

We would like to thank people from the Nasal and Besemah communities who took part in these documentation projects, especially Anton Supriyadi, Sarkani, Asfan Fikri Sanaf, and Kencana Dewi as well as collaborators on the Nasal project, including Yanti and Jacob Hakim. We are also grateful to the Ministry of Research and Technology in Indonesia for providing permissions for research on Besemah and Nasal as well as the Center for Culture and Language Studies at Atma Jaya Catholic University of Indonesia for sponsoring this research. This material is based upon work supported by the National Science Foundation under Grant No.~(1911641). Any opinions, findings, and conclusions or recommendations expressed in this material are those of the author and do not necessarily reflect the views of the National Science Foundation.

\bibliography{anthology,custom}
\bibliographystyle{acl_natbib}




\end{document}